\documentclass[journal,10pt,doublecolumn]{IEEEtran}

\usepackage{euscript}
\usepackage{amsmath}
\usepackage{amsthm}
\usepackage{amssymb}
\usepackage{epsfig}
\usepackage{xspace}
\usepackage{graphicx}
\usepackage{subcaption}
\usepackage{epstopdf}
\usepackage{epsfig}
\usepackage{color}
\usepackage{url}
\graphicspath{{Images/}}
\usepackage{multirow}
\usepackage{cite}
 \usepackage[table,xcdraw]{xcolor}
\usepackage{overpic}		
\usepackage{enumitem,kantlipsum}
\usepackage[ruled,vlined,linesnumbered]{algorithm2e}

\newcommand{\suchthat}{\;\ifnum\currentgrouptype=16 \middle\fi|\;}

\usepackage{enumitem}
\setlist[enumerate]{label*=.\arabic*, after=\normalfont}
\setlist[enumerate,1]{label=\arabic*, font=\normalfont\color{red}}
\setlist[enumerate,2]{before=\normalfont}
\setlist[enumerate,3]{font=\normalfont\upshape, before=\normalfont\itshape}

\usepackage{tikz}
\usepackage{collcell}

\begin{document}
\bstctlcite{IEEEexample:BSTcontrol}
\title{Source Printer Identification \\from Document Images Acquired using Smartphone}

\author{Sharad Joshi, Suraj Saxena, Nitin Khanna,~\IEEEmembership{Member,~IEEE}
\thanks{Sharad Joshi, Suraj Saxena, and Nitin Khanna are with the Multimedia Analysis and Security (MANAS) Lab, Electrical Engineering, Indian Institute of Technology Gandhinagar (IITGN), Gujarat, 382355 India. E-mail: \{nitinkhanna\}@iitgn.ac.in}}

\maketitle
\thispagestyle{empty}

\begin{abstract}
Vast volumes of printed documents continue to be used for various important as well as trivial applications. 
Such applications often rely on the information provided in the form of printed text documents whose integrity verification poses a challenge due to time constraints and lack of resources.
Source printer identification provides essential information about the origin and integrity of a printed document in a fast and cost-effective manner.
Even when fraudulent documents are identified, information about their origin can help stop future frauds.
If a smartphone camera replaces scanner for the document acquisition process, document forensics would be more economical, user-friendly, and even faster in many applications where remote and distributed analysis is beneficial.
Building on existing methods, we propose to learn a single CNN model from the fusion of letter images and their printer-specific noise residuals.
In the absence of any publicly available dataset, we created a new dataset consisting of 2250 document images of text documents printed by eighteen printers and acquired by a smartphone camera at five acquisition settings.
The proposed method achieves 98.42\% document classification accuracy using images of letter `e' under a 5$\times$2 cross-validation approach.
Further, when tested using about half a million letters of all types,
it achieves 90.33\% and 98.01\% letter and document classification accuracies, respectively, thus highlighting the ability to learn a discriminative model without dependence on a single letter type.
Also, classification accuracies are encouraging under various acquisition settings, including low illumination and change in angle between the document and camera planes.
\end{abstract}
\section{Introduction}
\IEEEPARstart{U}{sage} of digital documents has increased sharply in the last decade.
However, security issues, cost of transition, and acceptability by the workforce restrict a complete transition from printed to digital documents.
Such restrictions have encouraged continued usage of printed documents in many financial and administrative dealings such as agreements, deeds, business communication, and record-keeping.
So, there is co-existence of digital and printed documents.
As per a global forest product facts and figures 2018 report provided by the Food and Agriculture Organization of the United Nations, production of printing and writing paper was 96 million tonnes in 2018~\cite{web:paperusage2018} and has been steady since 2014.
The humongous volume of printed documents requires fast and accurate digital systems to predict their origin and integrity.
Information about the source printer can provide useful information about the origin and integrity of a printed document to an investigator.

The problem of attributing the source printer to a printed document has been studied extensively in the literature using digital methods~\cite{chiang2009printer,ferreira2017data}.
Two main approaches to source printer identification include (1) extrinsic methods based on embedding a signal in the printed document~\cite{Chiang2011} and (2) intrinsic methods that exploit artifacts introduced by the combination of various electro-mechanical parts of a printer~\cite{mikkilineni2011forensic}.
Apart from being costly and complex, extrinsic solutions require access to the printer before the document is printed, which may not be feasible as manufacturers are not legally bound to integrate such solutions into their printers.
On the other hand, intrinsic solutions only require sample document(s) printed from the printer and
can be used to investigate documents printed in the past.

\begin{figure}[t!]
    \centering
    \includegraphics[width=0.49\textwidth]{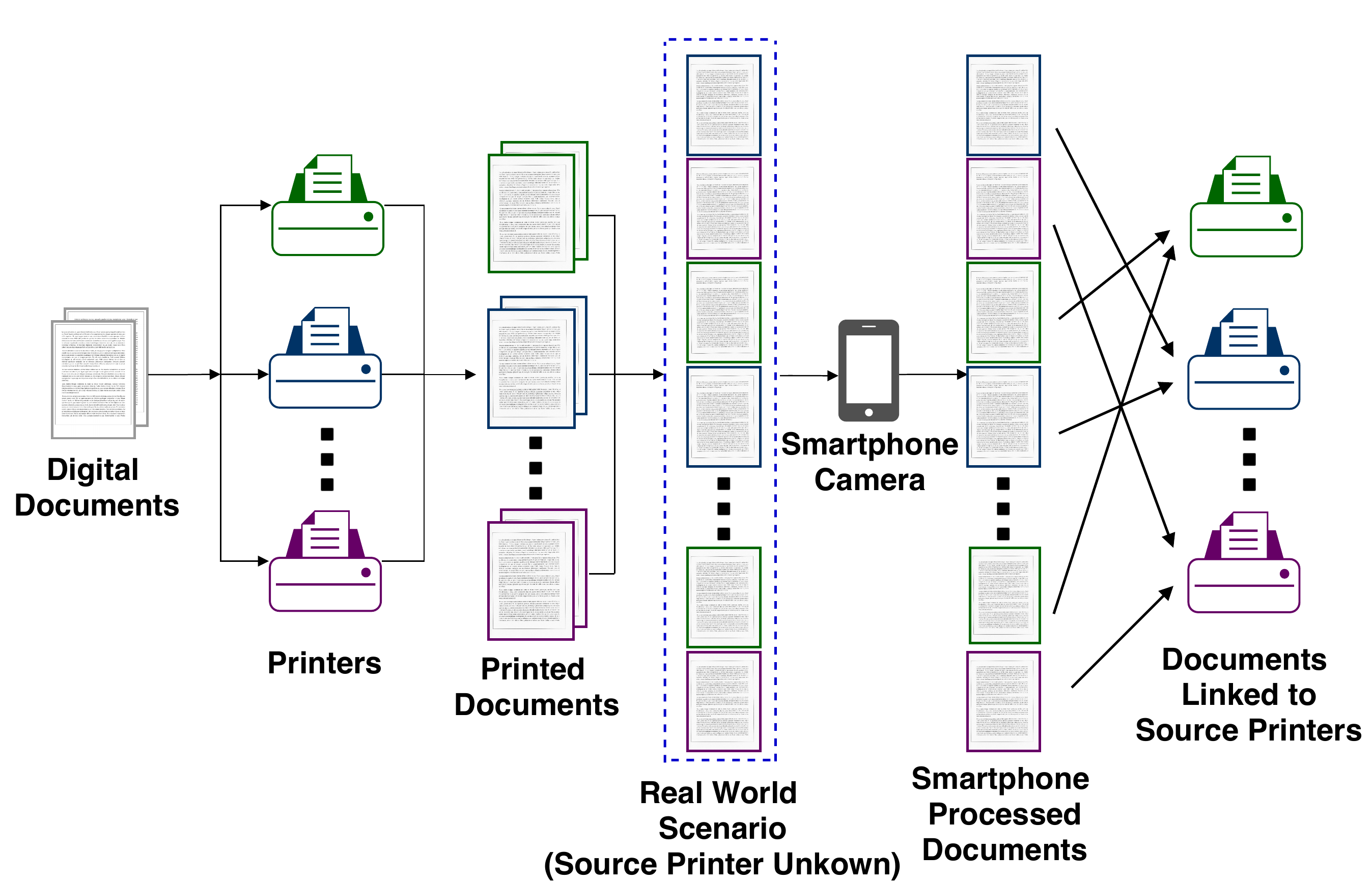}
    \caption{Overview of problem scenario}
    \label{fig:DeepPhone_Concept}
\end{figure}
All the existing methods in the literature make use of a reference scanner to acquire a digital image of the printed document.
On the other hand, smartphones with an in-built camera have become very common.
As compared to scanners, smartphones are compact, easy to use, and can be quickly deployed to acquire and transmit document images in a remote working environment.
Most importantly, they are light-weight gadgets that human beings have become accustomed to carry along almost all the time.
The document analysis community has recently started working towards replacing batch scanners by smartphone cameras~\cite{burie2015icdar2015}.
In document forensics, a very recent approach proposed a method for source identification of colored images printed by color laser printers \cite{kim2019learning}.
There are significant differences between the working of color and `black-and-white' (grayscale) printers.
So, the method proposed in~\cite{kim2019learning} cannot be used on grayscale documents.
In this work, we explore the scenario whereby a smartphone camera can replace the reference scanner used to convert text documents printed by black toner into digital images for source identification (Figure~\ref{fig:DeepPhone_Concept}).

Traditional source printer identification methods relied on handcrafted features extracted from texture patterns~\cite{ferreira2015laser,joshi2018single,joshi2019source}.
Recent works have shown that the convolutional neural network (CNN) can replace the traditional pattern recognition pipeline for attributing the source printer of printed text documents~\cite{ferreira2017data,joshiIcassp2018}.
In this work, we propose a combination of native letter image and its noise residual representation into a two-channel image, which allows learning a single CNN model.
Our experimental analysis shows that this strategy works better than learning multiple CNNs in parallel, one for each type of noise residual proposed in~\cite{ferreira2017data}.
We use a series of experiments to show the efficacy of the proposed two-channel CNN-based approach.
The major contributions of this work include the following: 
\begin{itemize}
    \item {To the best of our knowledge, this is the first approach for source printer identification of printed text documents using a smartphone camera.}
    \item {Introduces a single CNN approach, using a two-channel combination of letter image and its noise residual, capable of learning a single model.}
    \item {Introduces a new smartphone acquired dataset consisting of 2250 images of text documents printed from eighteen printers under five different types of settings.}
    \item {Performs better than state-of-the-art methods achieving 98.42\%, 94.70\%, 98.12\%, 97.86\%, and 97.69\% page-level classification accuracy using a $5\times2$ cross-validation approach with dataset acquired at 0$^{\circ}$, +5$^{\circ}$, -5$^{\circ}$,  free-hand, and low illumination settings, respectively.}
    \item{Achieves 90.33\% letter-level and 98.01\% document-level classification accuracy when tested on about half million letter images; thus paving the way for using all letter types and intra-page forensic tasks.}
\end{itemize}

We discuss the existing methods for source printer identification in Section~\ref{sec:DeepPhone_Existing}.
The proposed system has been described in Section~\ref{sec:DeepPhone_Proposed}. Section~\ref{sec:DeepPhone_Results} discusses an evaluation of the proposed method using a new smartphone acquired dataset.
At last, we discuss the outcomes of the proposed work and draw out the conclusions in Section~\ref{sec:DeepPhone_Conclusion}.

\section{Related Work}
\label{sec:DeepPhone_Existing}
The problem of source printer identification is well defined in the literature and has been addressed in detail in the last decade~\cite{chiang2009printer,ferreira2017data}.
Imperfections in the printing process lead to artifacts in the printed document, which are invisible to the naked eye~\cite{chiang2009printer}.
These artifacts may be observed in the zoomed versions of printed letters (Figure~\ref{fig:Sample_letters}).
Early methods relied on chemical and microscopic analysis. They require an expert's supervision, are time-consuming, and may even damage the printed document under investigation~\cite{girard2013criminalistics}.
The whole process has the potential to be automated using digital methods; speed depends upon the availability of computer hardware and does not require an expert's supervision.
Since printers treat text and images differently, source printer identification from printed text and printed images need to be addressed separately.
Since this work deals with text documents, only text-based methods are discussed here.
For a review of image-based methods, please refer~\cite{ferreira2015laser,ferreira2017data}.
Apart from the methods for source printer identification, some methods have been proposed to classify the printing techniques, i.e., laser, inkjet, or photocopier~\cite{oliver2002use, Lampert2007, schulze2008evaluation, Schulze2009, schreyer2009intelligent, roy2010authentication, shang2014detecting}.
We classify the text-based methods into texture-based, noise residual-based, deep learning-based, and other methods as follows.

\subsection{Texture-based Methods}
Researchers at Purdue University observed the appearance of light and dark lines perpendicular to the direction of the paper movement inside the printer. They termed it as \textit{banding}, which acted as one of the earliest intrinsic signature used in a digital system for source printer identification~\cite{chiang2009printer}.
The efficacy of the method was shown on documents scanned at high resolution (2400 dpi).

Consequently, texture-based methods gained popularity as they provide encouraging performance on documents scanned at lower resolutions (600 dpi).
At a high level of abstraction, most of these methods can be seen as following the typical pattern recognition pipeline. Such methods use one or more hand-crafted features extracted from either all occurrences of a specific letter type (like `e') or non-overlapping rectangular blocks obtained by segmenting the printed text documents.
These features are used to learn a suitable classifier model, which predicts printer labels for each letter or a group of letters in the document under test. 
Majority voting on these labels predicts the printer for the whole document.
All occurrences of the letter `e’ were used in~\cite{Mikkilineni2005} to extract gray-level co-occurrence matrices (GLCM) followed by calculation of 22 statistical features.
They used a 5-nearest neighbor classifier to predict printer labels.
The authors also extended this technique for other classifiers, namely, support vector machine
(SVM)~\cite{mikkilineni2005printer}, and Euclidean distances~\cite{mikkilineni2011forensic}.
Tsai \textit{et al.}~\cite{Tsai2014} used a combination of discrete wavelet transform (DWT) and GLCM based features extracted from all occurrences of a specific Chinese character followed by classification using SVM.
The same authors extended this method to a decision-fusion model-based approach for source printer identification, which applies a feature selection approach on a variety of features, including local binary pattern (LBP), GLCM and DWT~\cite{tsai2018decision}.
Ferreira \textit{et al.}~\cite{ferreira2015laser} used statistical features extracted from GLCM extended in multiple directions and scales as well as convolutional texture gradient filter (CTGF) which is a new feature descriptor based on filtering textures with a specific gradient. These features are extracted from all occurrences of `e' followed by SVM as the classifier.

All the methods discussed so far, extract features from only a particular letter type (usually the most frequent letter type).
Authors in~\cite{joshi2018single} proposed a single-classifier-based system that used features based on local tetra patterns (LTrP).
The whole system is designed such that a single SVM model can be applied to all the letters printed on a test document.
Recently, the same authors introduced a printer-specific local texture descriptor (PSLTD)~\cite{joshi2019source} that can be extracted from all the letters on a test page.
In contrast to previous methods, this method provides promising performance when the font type of letters in the test document is not present in the train documents. Thus, acting as the first step towards a font-independent descriptor.
\begin{figure*}[htb!]
    \centering
    \includegraphics[width=\textwidth]{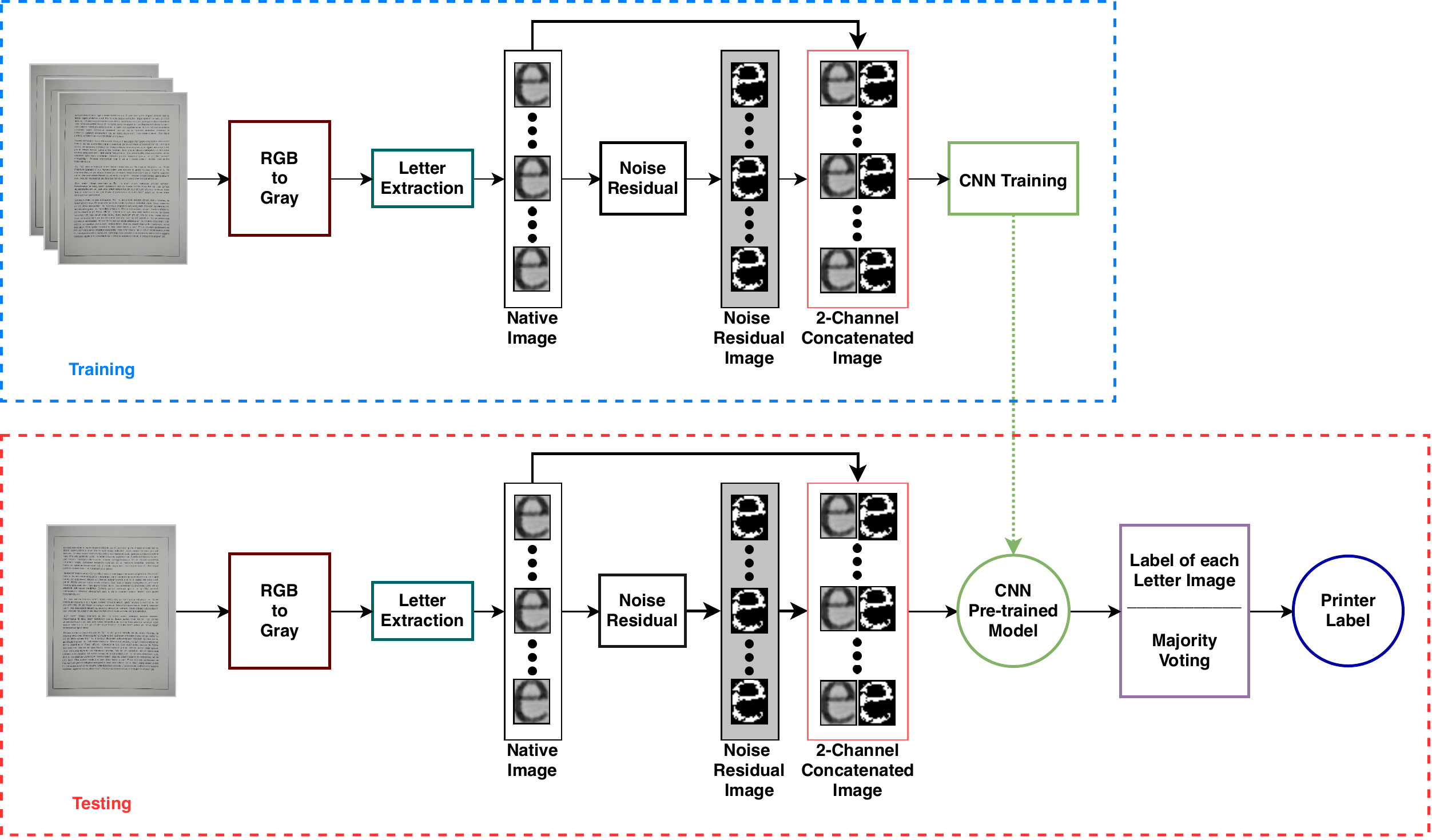}
    \caption{Overview of the proposed CNN based approach using printed documents captured by a smartphone camera.}
    \label{fig:DeepPhone_BD}
\end{figure*}

\subsection{Noise Residual-based Methods}
Some methods are based on extracting features from either only noise residual or a combination of native letter image and its noise residual.
Elkasrawi \textit{et al.} ~\cite{Elkasrawi2014} used statistical features extracted from the noise residual of each text line, followed by SVM as a classifier.
The method in~\cite{Tsai2014} is extended in~\cite{tsai2015japanese} by including more features extracted from noise residual obtained after applying a spatial filter, Wiener filter, and Gabor filter on native letter images (of a specific character type).

\subsection{Deep Learning-based Methods}
Ferreira \textit{et al.}~\cite{ferreira2017data} proposed a data-driven approach using six convolutional neural networks (CNN) trained in parallel.
Three CNNs learn models using all occurrences of letter `e', their mean filtered versions, and their median filtered versions.
The learned models are used to extract features that are concatenated and used as input to a one-vs-one SVM, which predicts printer labels for all images of letter `e'.
Similarly, a set of predictions are obtained for all images of letter `a'  using the other three CNNs and a one-vs-one SVM.
Majority voting on predicted letter labels provides printer labels for the whole document.

Authors in~\cite{joshiIcassp2018} proposed a noise residual which improves the performance using the CNN architecture utilized in~\cite{ferreira2017data}.
Also, they showed improvement in performance using printer-specific data-augmentation and a spatial pyramid pooling (SPP) layer~\cite{he2015spatial}.
The proposed method builds on this method and aims to reduce computational complexity, along with performance improvement.
However, data-augmentation and use of SPP layer significantly increase computation and memory requirements for a fixed size training data~\cite{joshi2019source}.
So, we do not consider the data-augmentation and SPP layer in the proposed method.

\subsection{Other Methods}
Kee \textit{et al.}~\cite{kee2008printer} applied principal component analysis (PCA) on all occurrences of `e'. They used the mean letter image and top p eigenvectors to characterize a printer profile.
Zhou \textit{et al.}~\cite{zhou2015text} proposed a text-independent approach using a specially designed and patented equipment to scan subtle textures on printed pages.
On the other hand, all other methods discussed here work with standard office scanners.
Another category of methods based on printer-specific geometric distortion has also been discussed in the literature~\cite{bulan2009geometric,wu2009printer,jain2019passive}.
Though promising, these methods require a reference soft copy of the printed document and document image scanned at high resolution~(1200 dpi).

\section{Proposed Method}
\label{sec:DeepPhone_Proposed}

\begin{figure}[t]
	\centering
	\begin{overpic}[width=0.25in]{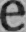}
	\end{overpic}\hspace{0.15in}
	\begin{overpic}[width=0.25in]{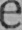}
	\end{overpic}\hspace{0.15in}
	\begin{overpic}[width=0.25in]{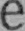}
	\end{overpic}\hspace{0.15in}
	\begin{overpic}[width=0.25in]{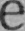}
	\end{overpic}\hspace{0.15in}
	\begin{overpic}[width=0.25in]{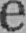}
	\end{overpic}\hspace{0.15in}
	\begin{overpic}[width=0.25in]{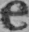}
	\end{overpic}\hspace{0.15in}
	\begin{overpic}[width=0.25in]{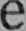}
	\end{overpic}\hspace{0.15in}
	\begin{overpic}[width=0.25in]{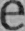}
	\end{overpic}\\
	\vspace{0.05in}
	\begin{overpic}[width=0.25in]{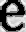}
	\end{overpic}\hspace{0.15in}
	\begin{overpic}[width=0.25in]{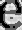}
	\end{overpic}\hspace{0.15in}
	\begin{overpic}[width=0.25in]{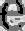}
	\end{overpic}\hspace{0.15in}
	\begin{overpic}[width=0.25in]{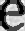}
	\end{overpic}\hspace{0.15in}
	\begin{overpic}[width=0.25in]{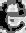}
	\end{overpic}\hspace{0.15in}
	\begin{overpic}[width=0.25in]{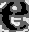}
	\end{overpic}\hspace{0.15in}
	\begin{overpic}[width=0.25in]{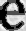}
	\end{overpic}\hspace{0.15in}
	\begin{overpic}[width=0.25in]{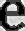}
	\end{overpic}\\
	\vspace{0.05in}
	\begin{overpic}[width=0.25in]{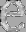}
	\end{overpic}\hspace{0.15in}
	\begin{overpic}[width=0.25in]{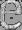}
	\end{overpic}\hspace{0.15in}
	\begin{overpic}[width=0.25in]{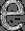}
	\end{overpic}\hspace{0.15in}
	\begin{overpic}[width=0.25in]{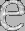}
	\end{overpic}\hspace{0.15in}
	\begin{overpic}[width=0.25in]{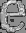}
	\end{overpic}\hspace{0.15in}
	\begin{overpic}[width=0.25in]{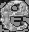}
	\end{overpic}\hspace{0.15in}
	\begin{overpic}[width=0.25in]{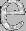}
	\end{overpic}\hspace{0.15in}
	\begin{overpic}[width=0.25in]{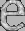}
	\end{overpic}\\
	\vspace{0.05in}
	\caption{Zoomed versions of sample letter images printed from 8 printers in our camera-acquired dataset: native letter ($1^{st}$ row), ideal letter ($2^{nd}$ row), and noise residual ($3^{rd}$ row).}
	\label{fig:Sample_letters}
\end{figure}

\begin{figure*}[htb!]
    \centering
    \includegraphics[width=\textwidth]{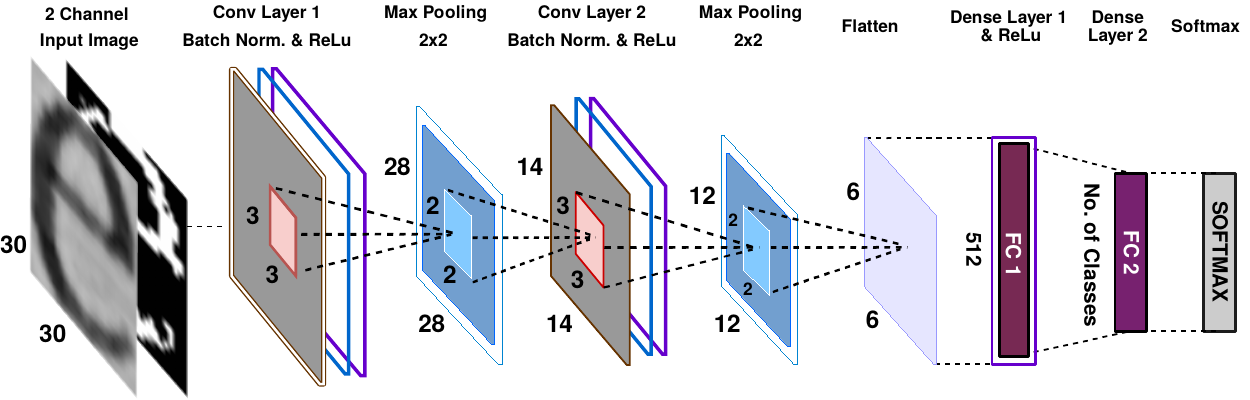}
    \caption{Overview of proposed CNN architecture.}
    \label{fig:DeepPhone_CNNArch}
\end{figure*}

In this paper, we propose to make use of a single CNN based method to learn a model that could find the source printer of printed documents using camera-acquired images of those documents.
The proposed method builds on the findings of~\cite{ferreira2017data}~and~\cite{joshiIcassp2018}.
Figure~\ref{fig:DeepPhone_BD} depicts the overview of the proposed method.
First module in the proposed method processes the document image to obtain two-channel letter images suitable for learning a printer discriminating CNN model. 
Then, an appropriate CNN architecture is designed. Finally, printer labels are predicted for a given test document.

\subsection{Pre-processing}
\label{subsec:DeepPhone_Proposed_preProcess}
At first, a reference smartphone is used to acquire images of all the printed text documents.
The default setting of most smartphones save the acquired image as a .jpg file, which is a three-channel RGB image.
Though, it would be worth exploring the impact of using an RGB document image instead of its gray-scale version. To reduce complexity, we assume that the black toner-based printer artifacts of our interest would be conveniently acquired in the gray-scale version.
So, in this work, we convert all RGB documents into their gray-scale versions.

All occurrences of letter `e' printed on a page are extracted using Matlab's in-built optical character recognition (OCR) in the form of a a rectangular image tightly bound around the printed letter (first row of Figure~\ref{fig:DeepPhone_BD}).
The hypothesis is that the printing process introduces certain device-specific artifacts, which can be described by the difference between the printed letter image and its \textit{ideally} printed version (termed here as the noise residual).
Also, the native letter and noise residual images have been shown to carry complementary information (Table 1 of~\cite{joshiIcassp2018}).
So, we estimate a noise residual image corresponding to each extracted letter image.
An \textit{ideal} printed letter image without such artifacts is expected to have equal toner spread in the interior region (i.e., a constant pixel intensity), and the region containing the white background must not contain any black toner.
However, zoomed letter images (Figure~\ref{fig:Sample_letters}) show that the black toner spreads at the border of a letter.
Thus, differentiation between a letter's interior region and its background is challenging.

In the absence of any other reliable source of information, we estimate the noise introduced during the printing and image-acquisition process.
We estimate an ideally printed letter image as described in~\cite{joshi2018single}.
First, we segment the letter image ($I$) into three regions, i.e., flat ($F$), edge ($E$), and background ($B$) using its intensity histogram as follows~\cite{joshi2018single,joshiIcassp2018}:
\begin{equation}
    \label{eqn:ErrResid_1_1}
    F = 
    \{ I(p) \suchthat I(p) \in[0,\alpha\mu ] \ \ and \ \ \forall p\in I \}
\end{equation}
\begin{equation}
    \label{eqn:ErrResid_1_2}
    E = 
    \{ I(p) \suchthat I(p) \in(\alpha\mu,\beta\mu] \ \ and \ \ \forall p\in I \}
\end{equation}
\begin{equation}
    \label{eqn:ErrResid_1_3}
    B = 
    \{ I(p) \suchthat I(p) \in(\beta\mu,max(I)] \ \ and \ \ \forall p\in I \}
\end{equation}
Here, $max(I)$ denote the maximum intensity value in $I$.
$\mu$ denotes the mean of all pixel intensities in $I$. $\alpha$ and $\beta$ are experimentally chosen constants which are fixed for all images across experiments.
Next, a three-level ideal image ($D$) is estimated from the input image as follows~\cite{joshiIcassp2018}: 
\begin{equation}
    \label{eqn:ErrResid_2_1}
    \small
    D(p) =  
    \begin{cases}
        median(F), &  I(p) \in(0,\alpha\mu] \\  
        median(E), & I(p) \in(\alpha\mu,\beta\mu] \\ 
        median(B), & I(p) \in(\beta\mu,max(I)] \\ 
    \end{cases}
\end{equation}
The native letter image ($I$) is subtracted from the ideal image (D) to obtain the noise residual image (N). 
Figure~\ref{fig:Sample_letters} shows samples of native letter images along with corresponding ideal and noise residual images.
Similar to the approach in~\cite{ferreira2017data}, both images are converted into two square images of fixed dimensions given by the patch size (hyperparameter).
Images having a larger dimension are center-cropped, i.e., extra rows or columns are removed from either side, whereas shorter dimensions are padded by zeros.
Our preliminary experiments suggested that the choice of padding by either zeros or the maximum possible pixel (i.e., 255 for an 8-bit image) did not have any significant impact on the classification accuracy.
However, since the pixel values in noise residual images are valued as `0' when there is no error, we chose padding by zeros.
Further, the letter images and their noise residuals are combined into 2-channel images such that CNN can learn a single model from them.

\subsection{Model Learning}
\label{subsec:DeepPhone_Proposed_model}

\begin{figure}[t!]
    \centering
        \includegraphics[width=0.5\textwidth]{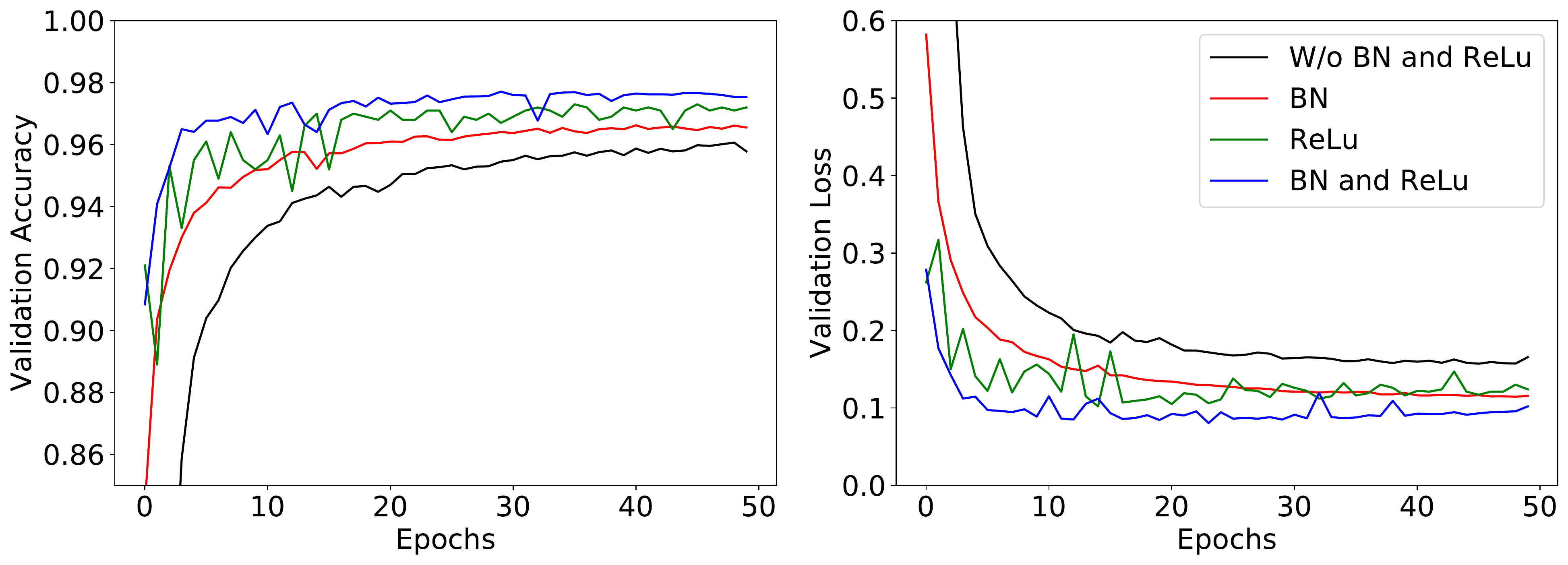}
    \caption{Performance with BN and ReLu by validation accuracy and validation loss (experimental settings discussed in Section~\ref{subsec:DeepPhone_dataset}).}
    \label{fig:DeepPhone_BNReLU}
\end{figure}

\begin{figure}[t!]
    \centering
    \begin{subfigure}[b]{0.48\columnwidth}
        \centering
        \includegraphics[width=\linewidth]{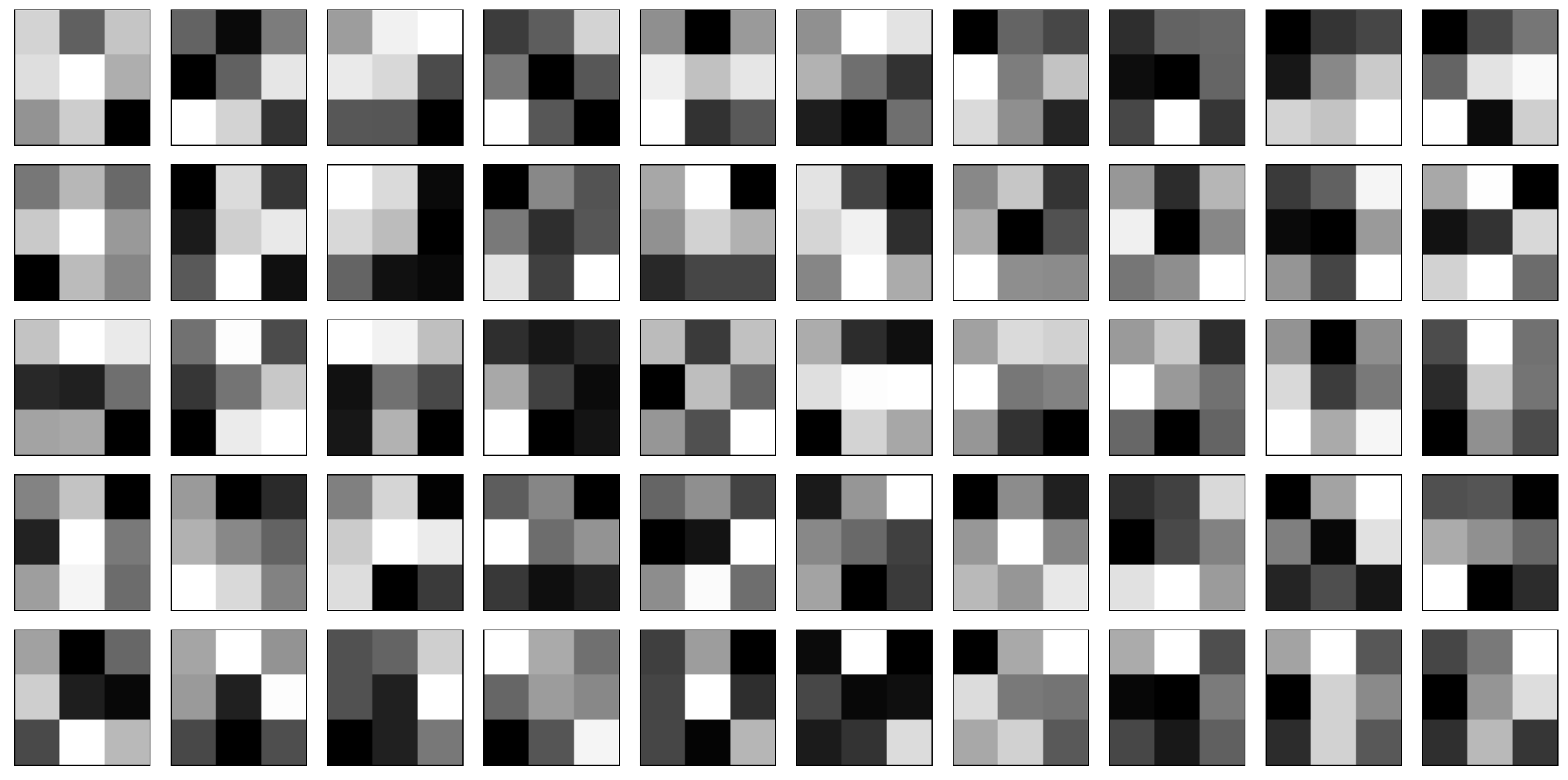}
        \caption{Camera Acquired (Native)}
        \label{fig:DeepPhone_filters_camNative}
    \end{subfigure}
    \begin{subfigure}[b]{0.48\columnwidth}
        \centering
        \includegraphics[width=\linewidth]{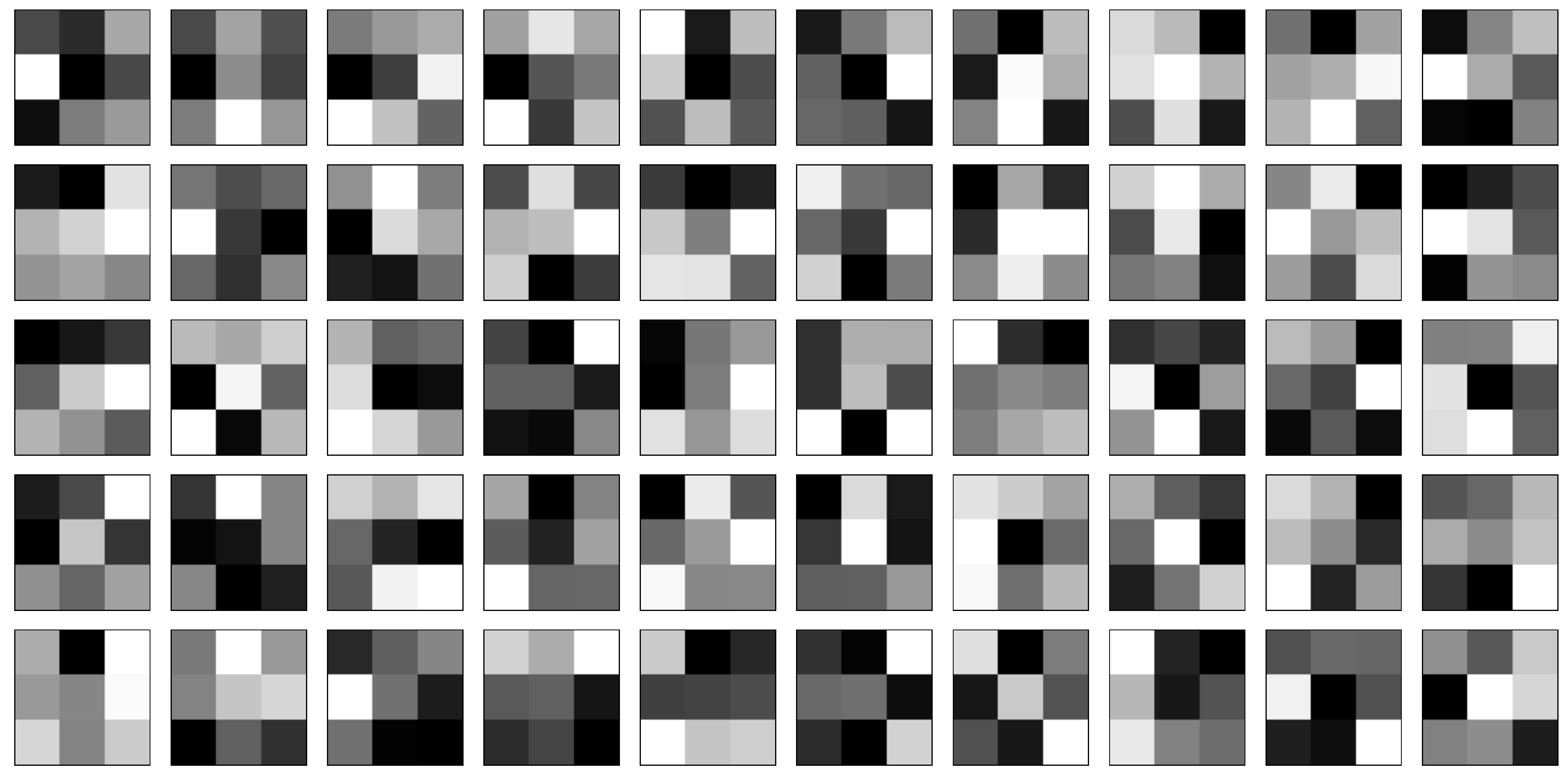}
        \caption{Camera Acquired (NR)}
        \label{fig:DeepPhone_filters_camNR}
    \end{subfigure}\\
    \vspace{0.1cm}
     \begin{subfigure}[b]{0.48\columnwidth}
        \centering
        \includegraphics[width=\linewidth]{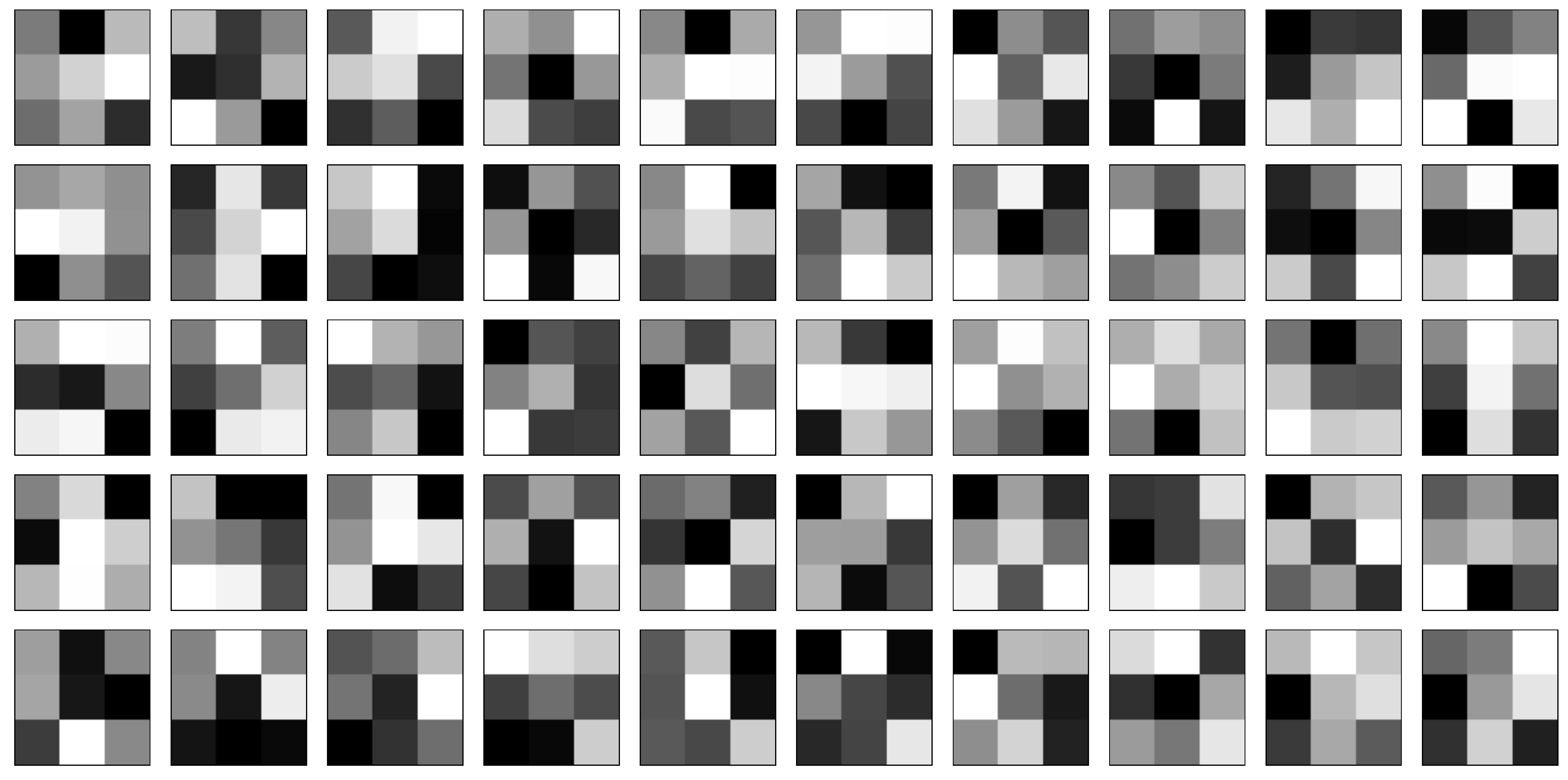}
        \caption{Scanner Acquired (Native)}
        \label{fig:DeepPhone_filters_scanNative}
    \end{subfigure}
     \begin{subfigure}[b]{0.48\columnwidth}
        \centering
        \includegraphics[width=\linewidth]{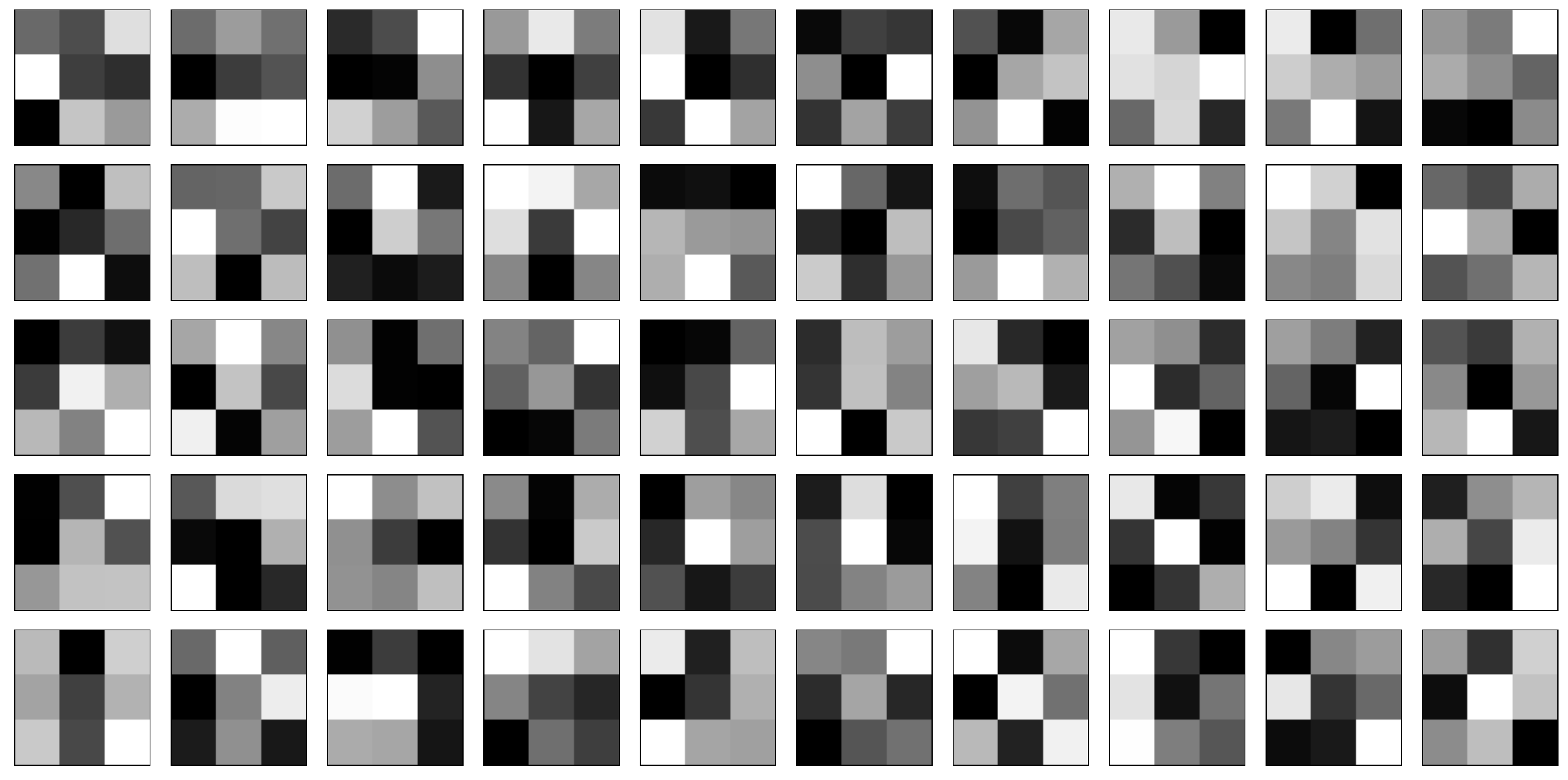}
        \caption{Scanner Acquired (NR)}
        \label{fig:DeepPhone_filters_scanNR}
    \end{subfigure}\\
    \caption{Visualization of filters (of first convolutional layer) learned from camera and scanner acquired train data (experimental settings discussed in Section~\ref{subsec:DeepPhone_dataset}; on one sample train and test fold).}
    \label{fig:DeepPhone_filters}
\end{figure}

The 2-channel letter images are used to train a single CNN model.
The CNN architecture used in~\cite{ferreira2017data} has been adapted for this problem as it has been shown to work well with small patches of letter images performing better than AlexNet~\cite{krizhevsky2012imagenet} and GoogLeNet~\cite{szegedy2015going}.
It is adapted to include a batch normalization (BN) layer~\cite{ioffe2015batch} to improve the speed of learning and an activation layer (ReLu) to introduce non-linearity that leads to improvement in classification accuracy by allowing the network to learn more discriminating models.
While designing the CNN architecture, we found that a combination of BN and ReLu performed better than using only either of them (Figure~\ref{fig:DeepPhone_BNReLU}).
Unlike document images acquired by a scanner, the information contained in camera acquired images may be corrupted due to lower resolution, uneven lighting (even with on-camera flash~\cite{fisher2001digital}), blur, and perspective distortion~\cite{liang2005camera}.
Further, the printer forensic traces may also be corrupted due to camera processing steps, which are optimized for general scene images and not for document forensic analysis.
These include sensor array (which induces sensor noise), demosaicing, and compression~\cite{MemonBook2013}.
To overcome such unwanted noise, our architecture consists of a larger number of smaller sized filters (i.e., 3 $\times$ 3) in each convolutional layer which can help extract a larger variety of filters from smaller image regions (filters are depicted in Figure~\ref{fig:DeepPhone_filters}).
The details of the CNN architecture are reported in~Table~\ref{tab:DeepPhone_CNNarch}.
Batch size is fixed at 100 whereas, Adam optimizer~\cite{kingma2014adam} is used with a learning rate of 0.001, decay of 0.0005 and cross-entropy loss.
The CNN model is trained for 50 epochs, and the model with the lowest validation loss is selected.

We visualize the 50 filters learned by the first convolutional layer in Figure~\ref{fig:DeepPhone_filters}.
Since the input to the CNN is a two-channel image, the first layer has 50 filters of dimensions 3 $\times$ 3 $\times$ 2, i.e., a total of 100 filters (50 $\times$ 2).
We show each filter channel separately arranged as a 5 $\times$ 10 matrix structure.
The model learns a variety of filters using both camera and scanner acquired images.
Most filters learned corresponding to native letter images acquired by smartphone camera are structurally similar to those learned from their scanner acquired version, with some variations in relative values.
This shows that the proposed method learns similar features from both types of data, thus showing the suitability of replacing a scanner with camera for document image-acquisition.
We emphasize that the models learned using both camera and scanner acquired document images contain the same printed text.

\begin{table*}[t!]
	\centering
	\caption{Details of proposed CNN architecture}
	\label{tab:DeepPhone_CNNarch}
\begin{tabular}{|c|c|c|c|c|c|}
\hline
\textbf{Layer}    & \textbf{Input Shape} & \textbf{Kernel Size} & \textbf{Stride} & \textbf{No. Filters} & \textbf{Output Shape} \\ \hline
Conv-1            & 30 x 30 x 2          & 3 x 3                & 1               & 50                   & 28 x 28 x 50          \\ \hline
Batch-Norm-1      & 28 x 28 x 50         & -                    & -               & -                    & 28 x 28 x 50          \\ \hline
ReLu Activation-1 & 28 x 28 x 50         & -                    & -               & -                    & 28 x 28 x 50          \\ \hline
Max Pooling-1     & 28 x 28 x 50         & 2 x 2                & 1               & -                    & 14 x 14 x 50          \\ \hline
Conv-2            & 14 x 14 x 50         & 3 x 3                & 1               & 50                   & 12 x 12 x 50          \\ \hline
Batch-Norm-2      & 12 x 12 x 50         & -                    & -               & -                    & 12 x 12 x 50          \\ \hline
ReLu Activation-2 & 12 x 12 x 50         & -                    & -               & -                    & 12 x 12 x 50          \\ \hline
Max Pooling-2     & 12 x 12 x 50         & 2 x 2                & 1               & -                    & 6 x 6 x 50            \\ \hline
Flatten           & 6 x 6 x 50           & -                    & -               & -                    & 1800                  \\ \hline
Dense Layer-1     & 1800                 & -                    & -               & -                    & 512                   \\ \hline
ReLu Activation-3 & 512                  & -                    & -               & -                    & 512                   \\ \hline
Dense Layer-2     & 512                  & -                    & -               & No. of Classes       & No. of Classes        \\ \hline
Softmax           & No. of Classes       & -                    & -               & -                    & No. of Classes        \\ \hline
\end{tabular}
\end{table*}

\subsection{Prediction}
A printed document under test goes through the pre-processing steps discussed in Section~\ref{subsec:DeepPhone_Proposed_preProcess}.
Then the CNN model learned in Section~\ref{subsec:DeepPhone_Proposed_model} using train data is used to predict the source printer labels of all letters (or all occurrences of a specific letter type) printed on the document.
A majority voting over the predicted labels of all letters printed on the document provides its source printer label.
\section{Experimental Evaluation}
\label{sec:DeepPhone_Results}
Since there is no publicly available dataset of printed text documents acquired by a smartphone camera, we created a new dataset for this purpose.
The proposed system has been extensively evaluated using a series of experiments.
We have also compared with several baseline methods including the state-of-the-art data driven system \emph{CNN-}$\{S^{raw}, S^{med}, S^{avg}\}_{a,e}$~\cite{ferreira2017data}, single classifier method \text{\emph{CC-RS-LTrP-PoEP}}~\cite{joshi2018single}, PSLTD~\cite{joshi2019source}, and three variations of DenseNet architecture~\cite{DenseCNN2017}.

\subsection{Dataset Details and Experimental Settings}
\label{subsec:DeepPhone_dataset}
We prepared a dataset by acquiring a total of 2250 images of text documents from Moto G3 smartphone with five different types of acquisition settings.
There are two types of variations considered in acquisition settings: (a) 
variation in relative angle between the camera and document planes (parallel (0$^{\circ}$), up-tilt (+5$^{\circ}$), down-tilt (-5$^{\circ}$), and free-hand)
and (b) change in illumination (low illumination).
It is worth noting here that the relative angle of 0$^{\circ}$ may not necessarily provide the same perspective in the document image as that acquired by a scanner.
This perspective distortion makes the characters printed farther from the camera look smaller, and some printed lines may not look parallel to each other~\cite{liang2005camera}.
Eighteen printers listed in Table I of~\cite{joshi2018single} printed twenty-five pages containing random text in Cambria font (English language) on white A-4 sheets using black colored toner.
These are the same documents used to create our scanner acquired
dataset~\cite{joshi2018single} (for more details about the printed documents, please refer~\cite{joshi2018single}~and~\cite{joshi2019source}).
Their visual appearance is grayscale, but, when a smartphone acquires images, it saves them as three-channel RGB Images.
The smartphone acquires each page only once (with a particular acquisition setting) using the default smartphone setting with HDR mode set to off. 
Except for the free-hand scenario, the smartphone acquires document images in a fixed indoor setup, where we fixed the location of the smartphone and documents using an appropriate physical structure. Specifically, we designated an area to place each document (i.e., a printed page) to be acquired and placed a selfie stick on top of it.
The selfie stick held our reference smartphone such that the angle between document and smartphone could be adjusted and fixed as per our requirements. The setup allowed us to acquire multiple versions of the dataset at an angle of 0, 5, and -5 degrees.
We created this setup to reduce the amount of human mishandling error and to analyze the impact of acquisition settings in a better manner.
We also acquired images in low lighting conditions at an angle of 0 degrees to analyze its impact on classification accuracy.

In contrast, the free-hand setting mimics a real-world scenario where a ``common'' camera-user tries to adjust the camera such that the document image is of good perceptual quality. So, in this scenario, there are no restrictions on the specific placement of the phone relative to a document.
Also, the relative angle between the camera and document planes may not be consistent over the acquisition of multiple printed documents.

Unless stated otherwise, all experimental results have been carried out using a 5 $\times$ 2 cross-validation approach on our dataset acquired at 0 degrees. For the first five-folds, out of 25 documents per printer, 12 printed documents are used in training and 13 in testing.
Similarly, for the rest of the five folds, train and test data are exchanged, such that there are 13 documents in training and 12 in testing.
Thus, there are a total of 450 document images acquired for each setting and we split them almost equally into train and test for each fold.
Each document contains about 2300 letters (about 300 occurrences of letter `e') and there are about 1 million letters overall.
The train and test sets are disjoint for each fold.
Moreover, the classification accuracy stats reported here are mean, median, and standard deviation ($\sigma$) values of accuracies obtained over all ten folds.

\begin{table}[t!]
	\centering
	\caption{Effect of patch size: classification accuracy statistics using a 5 $\times$ 2 cross-validation approach}
	\label{tab:DeepPhone_PatchSize}
	\resizebox{0.48\textwidth}{!}{
\begin{tabular}{|c|c|c|c|c|c|c|}
\hline
\textbf{Patch Size} & \textbf{22} & \textbf{24} & \textbf{26} & \textbf{28} & \textbf{30}    & \textbf{32}    \\ \hline
\textbf{Mean} (in \%)       & 97.97       & 97.70       & 98.01       & 98.28       & \textbf{98.42} & 98.37          \\ \hline
\textbf{Median} (in \%)     & 98.01       & 97.77       & 98.01       & 98.22       & \textbf{98.29} & \textbf{98.29} \\ \hline
\textbf{$\sigma$}        & \textbf{0.59}        & 0.75        & 0.67        & 0.69        & 0.66           & 0.69           \\ \hline
\end{tabular}
}
\end{table}

\begin{table}[t!]
	\centering
	\caption{Comparison of various approaches: classification accuracy statistics (mean and median accuracies in \%) and standard deviation in brackets using 5 iterations.}
	\label{tab:DeepPhone_approaches}
	\resizebox{0.50\textwidth}{!}{
\begin{tabular}{|c|c|c|c|c|c|c|c|c|}
\hline
\multirow{2}{*}{\textbf{Method}}                                           & \multirow{2}{*}{\textbf{Stats}}                             & \multicolumn{6}{c|}{\textbf{Train Pages}}                                                                                                                                                                                                                                                                                                                                                     & \multirow{2}{*}{\textbf{\begin{tabular}[c]{@{}c@{}}Model\\ Params\end{tabular}}} \\ \cline{3-8}
                                                                           &                                                             & \textbf{2}                                                    & \textbf{4}                                                    & \textbf{6}                                                    & \textbf{8}                                                    & \textbf{10}                                                   & \textbf{12}                                                   &                                                                                  \\ \hline
\textbf{Native}                                                            & \begin{tabular}[c]{@{}c@{}}Mean \\ Median\\ $\sigma$\end{tabular} & \begin{tabular}[c]{@{}c@{}}\textbf{94.44}\\ 94.87 \\ 1.60\end{tabular} & \begin{tabular}[c]{@{}c@{}}95.98\\ 96.58 \\ 1.74\end{tabular} & \begin{tabular}[c]{@{}c@{}}97.35\\ 97.44 \\ 0.68\end{tabular} & \begin{tabular}[c]{@{}c@{}}96.41\\ 97.01 \\ 2.17\end{tabular} & \begin{tabular}[c]{@{}c@{}}97.44\\ 97.44 \\ 0.81\end{tabular} & \begin{tabular}[c]{@{}c@{}}\textbf{98.21}\\ 98.29 \\ 0.50\end{tabular} & $\sim$955 k                                                                       \\ \hline
\textbf{NR}                                                                & \begin{tabular}[c]{@{}c@{}}Mean\\ Median\\ $\sigma$\end{tabular} & \begin{tabular}[c]{@{}c@{}}93.08\\ 93.16 \\ 1.16\end{tabular} & \begin{tabular}[c]{@{}c@{}}95.47\\ 94.87 \\ 1.23\end{tabular} & \begin{tabular}[c]{@{}c@{}}96.58\\ 96.58 \\ 0.81\end{tabular} & \begin{tabular}[c]{@{}c@{}}96.75\\ 97.01 \\ 0.84\end{tabular} & \begin{tabular}[c]{@{}c@{}}96.92\\ 96.58 \\ 0.63\end{tabular} & \begin{tabular}[c]{@{}c@{}}97.44\\ 97.44 \\ 0.72\end{tabular} & $\sim$955 k                                                                       \\ \hline
\textbf{\begin{tabular}[c]{@{}c@{}}2 Channel \\ (Native, NR)\end{tabular}} & \begin{tabular}[c]{@{}c@{}}Mean\\ Median\\ $\sigma$\end{tabular} & \begin{tabular}[c]{@{}c@{}}93.33\\ 94.02 \\ 1.39\end{tabular} & \begin{tabular}[c]{@{}c@{}}\textbf{96.58}\\ 96.58 \\ 0.97\end{tabular} & \begin{tabular}[c]{@{}c@{}}\textbf{97.44}\\ 97.44 \\ 0.60\end{tabular} & \begin{tabular}[c]{@{}c@{}}\textbf{96.67}\\ 97.44 \\ 1.99\end{tabular} & \begin{tabular}[c]{@{}c@{}}\textbf{97.52}\\ 97.44 \\ 0.68\end{tabular} & \begin{tabular}[c]{@{}c@{}}98.03\\ 98.29 \\ 0.51\end{tabular} & $\sim$955 k                                                                       \\ \hline
\textbf{\begin{tabular}[c]{@{}c@{}}Parallel \\ (Native, NR)\end{tabular}}  & \begin{tabular}[c]{@{}c@{}}Mean\\ Median\\ $\sigma$\end{tabular} & \begin{tabular}[c]{@{}c@{}}93.76\\ 93.59 \\ 1.30\end{tabular} & \begin{tabular}[c]{@{}c@{}}95.73\\ 95.51 \\ 0.36\end{tabular} & \begin{tabular}[c]{@{}c@{}}96.97\\ 97.01 \\ 0.71\end{tabular} & \begin{tabular}[c]{@{}c@{}}96.58\\ 96.79\\ 1.22\end{tabular}  & \begin{tabular}[c]{@{}c@{}}97.18\\ 97.01 \\ 0.71\end{tabular} & \begin{tabular}[c]{@{}c@{}}97.82\\ 97.86 \\ 0.56\end{tabular} & $\sim$1,910 k                                                 \\ \hline
\end{tabular}
}
\end{table}

\subsection{Effect of Patch Size}
\label{subsec:DeepPhone_patchsize}
First, we analyze the effect of different patch sizes on the performance of the proposed system.
For this purpose, we computed the median height and width of all occurrences of the letter `e' in our dataset acquired at 0 degrees, which are 31 and 26, respectively.
Based on this observation, we select the range of patch sizes from 22 to 32.
The results in Table~\ref{tab:DeepPhone_PatchSize} show that the proposed system performs well with patches of all sizes.
Moreover, patches of size 32 $\times$ 32 and 30 $\times$ 30 provide the highest average classification accuracy of 98.42\% when averaged over ten folds used in a 5 $\times$ 2 cross-validation approach.
However, the standard deviation~($\sigma$) of classification accuracies over the ten folds is smaller using the patch size as 30 $\times$ 30, i.e., 0.66.
So, for the rest of the experiments discussed in this section, we fix the patch size as 30 $\times$ 30 while experimenting with the proposed system.

\begin{table}[t!]
	\centering
	\caption{Comparison with state-of-the-art: classification accuracy statistics using 5 $\times$ 2 cross-validation approach.}
	\label{tab:DeepPhone_State-art}
	\resizebox{0.48\textwidth}{!}{
\begin{tabular}{|c|c|c|c|}
\hline
\textbf{Method}             & \textbf{Mean} (in \%)  & \textbf{Median} (in \%) & \textbf{$\sigma$}  \\ \hline
\textbf{\emph{CNN-}$\{S^{raw}, S^{med}, S^{avg}\}_{a,e}$~\cite{ferreira2017data}} & 96.58          & 97.01           & 2.32          \\ \hline
\textbf{\text{\emph{CC-RS-LTrP-PoEP}}~\cite{joshi2018single}}  & 93.99          & 94.23           & 1.37          \\ \hline
\textbf{PSLTD~\cite{joshi2019source}}              & 96.58          & 97.01           & 2.32          \\ \hline
\textbf{DenseNet121~\cite{DenseCNN2017}}        & 97.21          & 97.12           & 0.70          \\ \hline
\textbf{DenseNet169~\cite{DenseCNN2017}}        & 97.12          & 97.12           & 0.67          \\ \hline
\textbf{DenseNet201~\cite{DenseCNN2017}}        & 97.21          & 97.12           & 0.77          \\ \hline
\textbf{Proposed}           & \textbf{98.42} & \textbf{98.29}  & \textbf{0.66} \\ \hline
\end{tabular}
}
\end{table}

\begin{table*}[htb!]
	\centering
	\caption{Average confusion matrix (in \%) obtained using proposed method on dataset acquired using free-hand setting.}
	\label{tab:DeepPhone_free_Confmat}
	\resizebox{\textwidth}{!}{
	\begin{tabular}{|c|c|c|c|c|c|c|c|c|c|c|c|c|c|c|c|c|c|c|}
\hline
                                & \multicolumn{18}{c|}{\textbf{Predicted}}                                                                                                                                                                                                                                                                                                                                                                                                                                                                                                                                                                                                                                                                                              \\ \cline{2-19} 
\multirow{-2}{*}{\textbf{True}} & \textbf{B7065}                        & \textbf{CD52V}                        & \textbf{CI657}                        & \textbf{CIR50}                       & \textbf{CIR70}                        & \textbf{CIR85}                        & \textbf{CL29B}                       & \textbf{CL505}                       & \textbf{C432D}                        & \textbf{C482D}                       & \textbf{C4822}                       & \textbf{C4823}                       & \textbf{EL800}                        & \textbf{EL360}                        & \textbf{H1020}                        & \textbf{H1005}                       & \textbf{KM215}                       & \textbf{R5002}                        \\ \hline
\textbf{B7065}                  & \cellcolor[HTML]{67FD9A}\textbf{94.6} & \cellcolor[HTML]{FD6864}4.7           &                                       &                                      & \cellcolor[HTML]{FFFFC7}0.8           &                                       &                                      &                                      &                                       &                                      &                                      &                                      &                                       &                                       &                                       &                                      &                                      &                                       \\ \hline
\textbf{CD52V}                  & \cellcolor[HTML]{FFFE65}1.6           & \cellcolor[HTML]{67FD9A}\textbf{97.6} &                                       &                                      & \cellcolor[HTML]{FFFFC7}0.8           &                                       &                                      &                                      &                                       &                                      &                                      &                                      &                                       &                                       &                                       &                                      &                                      &                                       \\ \hline
\textbf{CI657}                  &                                       &                                       & \cellcolor[HTML]{67FD9A}\textbf{96.7} &                                      &                                       &                                       &                                      &                                      &                                       &                                      &                                      &                                      & \textbf{}                             & \cellcolor[HTML]{FFFE65}3.3           &                                       &                                      &                                      &                                       \\ \hline
\textbf{CIR50}                  &                                       &                                       &                                       & \cellcolor[HTML]{67FD9A}\textbf{100} &                                       &                                       &                                      &                                      &                                       &                                      &                                      &                                      &                                       &                                       &                                       &                                      &                                      &                                       \\ \hline
\textbf{CIR70}                  &                                       & \cellcolor[HTML]{FFFE65}2.4           &                                       &                                      & \cellcolor[HTML]{67FD9A}\textbf{97.6} &                                       &                                      &                                      &                                       &                                      &                                      &                                      &                                       &                                       &                                       &                                      &                                      &                                       \\ \hline
\textbf{CIR85}                  &                                       & \cellcolor[HTML]{FFFFC7}0.8           &                                       &                                      &                                       & \cellcolor[HTML]{67FD9A}\textbf{99.2} &                                      &                                      &                                       &                                      &                                      &                                      &                                       &                                       &                                       &                                      &                                      &                                       \\ \hline
\textbf{CL29B}                  &                                       &                                       &                                       &                                      &                                       &                                       & \cellcolor[HTML]{67FD9A}\textbf{100} &                                      &                                       &                                      &                                      &                                      &                                       &                                       &                                       &                                      &                                      &                                       \\ \hline
\textbf{CL505}                  &                                       &                                       &                                       &                                      &                                       &                                       &                                      & \cellcolor[HTML]{67FD9A}\textbf{100} &                                       &                                      &                                      &                                      &                                       &                                       &                                       &                                      &                                      &                                       \\ \hline
\textbf{C432D}                  &                                       &                                       & \cellcolor[HTML]{FFFE65}2.4           &                                      &                                       &                                       &                                      &                                      & \cellcolor[HTML]{67FD9A}\textbf{97.6} &                                      &                                      &                                      &                                       &                                       &                                       &                                      &                                      &                                       \\ \hline
\textbf{C482D}                  &                                       &                                       &                                       &                                      &                                       &                                       &                                      &                                      &                                       & \cellcolor[HTML]{67FD9A}\textbf{100} &                                      &                                      &                                       &                                       &                                       &                                      &                                      &                                       \\ \hline
\textbf{C4822}                  &                                       &                                       &                                       &                                      &                                       &                                       &                                      &                                      &                                       &                                      & \cellcolor[HTML]{67FD9A}\textbf{100} &                                      &                                       &                                       &                                       &                                      &                                      &                                       \\ \hline
\textbf{C4823}                  &                                       &                                       &                                       &                                      &                                       &                                       &                                      &                                      &                                       &                                      &                                      & \cellcolor[HTML]{67FD9A}\textbf{100} &                                       &                                       &                                       &                                      &                                      &                                       \\ \hline
\textbf{EL800}                  &                                       &                                       & \cellcolor[HTML]{FFFE65}2.4           &                                      &                                       &                                       &                                      &                                      &                                       &                                      &                                      &                                      & \cellcolor[HTML]{67FD9A}\textbf{94.3} & \cellcolor[HTML]{FFFE65}3.3           &                                       &                                      &                                      &                                       \\ \hline
\textbf{EL360}                  &                                       &                                       & \cellcolor[HTML]{FD6864}5.4           &                                      &                                       &                                       &                                      &                                      &                                       &                                      &                                      &                                      &                                       & \cellcolor[HTML]{67FD9A}\textbf{94.6} &                                       &                                      &                                      &                                       \\ \hline
\textbf{H1020}                  &                                       &                                       & \cellcolor[HTML]{FFFFC7}0.8           &                                      &                                       &                                       &                                      &                                      &                                       &                                      &                                      &                                      &                                       &                                       & \cellcolor[HTML]{67FD9A}\textbf{99.2} &                                      &                                      &                                       \\ \hline
\textbf{H1005}                  &                                       &                                       &                                       &                                      &                                       &                                       &                                      &                                      &                                       &                                      &                                      &                                      &                                       &                                       &                                       & \cellcolor[HTML]{67FD9A}\textbf{100} &                                      &                                       \\ \hline
\textbf{KM215}                  &                                       &                                       &                                       &                                      &                                       &                                       &                                      &                                      &                                       &                                      &                                      &                                      &                                       &                                       &                                       &                                      & \cellcolor[HTML]{67FD9A}\textbf{100} &                                       \\ \hline
\textbf{R5002}                  &                                       &                                       &                                       &                                      &                                       &                                       &                                      &                                      &                                       &                                      &                                      &                                      &                                       &                                       & \cellcolor[HTML]{FFFE65}2.3           &                                      &                                      & \cellcolor[HTML]{67FD9A}\textbf{97.7} \\ \hline
\end{tabular}
	}
\end{table*}

\subsection{Combination of Native and Noise Residual Images}
\label{subsec:DeepPhone_CNNprotocol}
We evaluate the impact of learning CNN models using only native letter images, only noise residual images, a combination of both, and two parallel CNNs from native and noise residual images.
As expected, in general, the classification accuracy increases with the number of training pages as shown in Table~\ref{tab:DeepPhone_approaches}.
All experiments have been carried out using five iterations with varying documents in train and test.
However, the five combinations are fixed across all methods for a given number of training pages.
The results show that two-channel concatenation of native and residual images works consistently better than using two CNNs in parallel with different number of training pages.
Moreover, learning a CNN model from only noise residual versions performs the worst among all variants.
We also compare the number of CNN model parameters used in each variant. 
The native, noise residual, and the two-channel approaches require almost the same, i.e., about 955,000 model parameters.
Wehereas, the two parallel CNNs-based approach requires almost double the number of model parameters i.e., about 1,910,000.

The results in Table I of~\cite{joshiIcassp2018} confirmed that using the scanner-acquired public dataset of~\cite{ferreira2015laser}, there is a consistent improvement in classification accuracies (for all the ten folds) by using a combination of native and noise residual versions as compared to using only native images.
But, the experiments with our camera-acquired dataset show that the proposed CNN architecture provides almost similar performance with only native letter images as well as with the two-channel combination of native images and their noise residual versions.
Specifically, the 2-channel approach works slightly better when training pages equal to 4, 6, 8 and 10 documents per printer.
Also, an evaluation of all types of letter images provides 98.01\% and 97.87\% average classification accuracy for the 2-channel and native only approaches, respectively.
One possible reason for this reduced impact of noise residual could be the extra noise introduced during the acquisition process by the camera as compared to a scanner.
Nonetheless, we use the two-channel approach while discussing the rest of the experiments highlighting the use of a camera in place of a scanner.

\subsection{Comparison with State-of-the-art Methods}
\label{subsec:DeepPhone_state-of-the-art}
We compare the proposed system with state-of-the-art methods as well as a popular state-of-the-art CNN architecture for image classification tasks, i.e., DenseNet~\cite{DenseCNN2017}.
All accuracy results are reported using a 5 $\times$ 2 cross-validation approach.
The results in Table~\ref{tab:DeepPhone_State-art} show that the proposed method outperforms existing methods.
Also, the proposed method performs better than three variations of the DenseNet architecture, i.e., DenseNet121, DenseNet169, and DenseNet201, which contain 121, 169, and 201 layers, respectively.
Apart from achieving the highest classification accuracy, the standard deviation is also the lowest with the proposed method, thus outlining its consistency in comparison to other methods.
For a fair comparison, the input to DenseNet is the two-channel image, which is the same as the input to the proposed CNN architecture.
The experiment on DenseNet was run using the in-built option provided by Keras~\cite{kerasapplication} using the minimum allowed patch size, i.e., 32 $\times$ 32.

\begin{table}[t!]
	\centering
	\caption{Effect of variations in acquisition settings: classification accuracy statistics using 5 $\times$ 2 cross-validation}
	\label{tab:DeepPhone_acquisition}
\begin{tabular}{|c|c|c|c|}
\hline
\textbf{Method}    & \textbf{Mean} (in \%) & \textbf{Median} (in \%) & \textbf{$\sigma$} \\ \hline
\textbf{Parallel (0$^\circ$)}  & \textbf{98.42}         & \textbf{98.29}           & \textbf{0.66}         \\ \hline
\textbf{Down-tilt (-5$^\circ$)} & 94.70         & 94.87           & 0.64         \\ \hline
\textbf{Up-tilt (+5$^\circ$)}   & 98.12         & 97.86           & 0.69         \\ \hline
\textbf{Free-hand} & 97.86         & 97.44           & 0.90         \\ \hline
\textbf{Low Illumination} & 97.69         & 97.86           & 1.03         \\ \hline
\end{tabular}
\end{table}
\subsection{Effect of Acquisition Settings on Classification Accuracy}
\label{subsec:DeepPhone_acquisitionsettings}
We evaluate the proposed system under five acquisition settings on the nature of the surroundings and the image acquisition technique used by an investigator.
The first variation pertains to the change in the relative angle between the camera and document planes.
Here, we assume three possibilities: (a) the camera is placed exactly parallel to the document, (b) camera is at some angle relative to the document, and (c) the angle is not fixed and may arbitrarily vary while capturing both train and test documents.
The second variation is related to the amount of illumination available at the time of capturing the images.
Here, we use a much lower illumination than the previous scenarios.
The results show that the proposed method works consistently well for all the above scenarios~(Table~\ref{tab:DeepPhone_acquisition}), including the free-hand scenario, which is the easiest to implement in a practical scenario without any constraints.
As expected, the average classification accuracy is highest using the controlled setting where camera and document planes are almost parallel to each other while it is lowest at 94.70\% for the down-tilt (-5$^{\circ}$) setting.
The confusion matrix for the free-hand scenario (Tables~\ref{tab:DeepPhone_free_Confmat}) shows that the proposed system works consistently for identifying the documents printed by all the printers in our dataset.
The proposed method correctly classifies all test documents printed by eight out of eighteen printers over all ten folds in the 5 $\times$ 2 cross-validation.

However, mis-classifications are higher using the down-tilt (-5$^{\circ}$) setting because the proposed system extracts fewer letters.
Specifically, about 18,486 occurrences of letter `e' are extracted from down-tilt (-5$^{\circ}$) setting as compared to 51,894 and 75,060 from up-tilt (5$^{\circ}$) and free-hand, respectively.
This is a limitation of the letter extraction stage which currently uses a default OCR.
The decrease in OCR accuracy is a result of increase in perspective distortion~\cite{liang2005camera} as well as misalignment of camera and document planes.
Flatbed scanners do not suffer from this limitation as the documents and scanning surface are perfectly aligned.
However, since the investigator can control the image acquisition process, such a scenario may be avoided even without the use of a more recent and improved letter extraction approach.
We emphasize that the same printed documents are used for training and testing in experiments with all the acquisition settings.


\begin{table}[t!]
\centering
    \caption{Experiment using all letter types: classification accuracy statistics using 5 $\times$ 2 cross-validation}
	\label{tab:DeepPhone_AllLetters}
\begin{tabular}{|c|c|c|c|}
\hline
                    & \textbf{Mean} (in \%) & \textbf{Median} (in \%) & \textbf{$\sigma$} \\ \hline
\textbf{Page Level} & 98.01                  & 98.22                    & 0.70                   \\ \hline
\textbf{Char Level} & 90.33                  & 90.20                    & 0.53                   \\ \hline
\end{tabular}
\end{table}
\subsection{Experiment using All Types of Letters}
\label{subsec:DeepPhone_all}
The single-classifier approach introduced in~\cite{joshi2018single} showed that it is possible to learn a single classifier model using all letter types.
This approach is particularly useful when a specific type of letters printed on the available documents is insufficient to train a discriminative classifier model, as shown in Table V of~\cite{joshi2018single}.
We analyze the ability of the proposed method to learn a single CNN model from all types of letters in Table~\ref{tab:DeepPhone_AllLetters}.
Results show that the proposed method can achieve performance similar to using only a single letter type (i.e., `e').
Moreover, the average character level accuracy on about half a million letters is 90.33\%.
Such a high accuracy highlights that the CNN model is attributing the artifacts in a vast majority of letter images to the source printer, thus further strengthening the argument of using a camera in place of a scanner.
Further, it helps pave the way for intra-page forensic tasks such as detection of forgery created by either printing in empty spaces or pasting a small printed chit on portions of a text document~\cite{van2013text}.

\subsection{Experiment on Scanner Dataset}
\label{subsec:DeepPhone_scanner}
We also analyze the performance of the proposed method on our scanner acquired dataset~\cite{joshi2019source}.
We use the experimental settings used to report the same-font and cross-font results in Table IV of~\cite{joshi2019source}.
The results (Table~\ref{tab:DeepPhone_Scanner}) show that the proposed method performs equivalent to the state-of-the-art methods for the same-font scenario (i.e., font type of letters in test data is present in train data) by correctly classifying test document images from all printers.
But, it fails with cross-font scenario (i.e., font type of letters in test data is absent in train data) similar to existing data-driven methods of~\cite{ferreira2017data}~and~\cite{joshiIcassp2018}.
It is expected as the proposed method is dependent upon CNN to learn appropriate features based on the training data.

\begin{table}[t!]
	\centering
	\caption{Performance of proposed method on dataset acquired by scanner: classification accuracies (in \%) in a cross-font scenario}
	\label{tab:DeepPhone_Scanner}
\begin{tabular}{|c|c|c|c|c|}
\hline
\textbf{Train Font}      & \multicolumn{4}{c|}{\textbf{Cambria (C)}}           \\ \hline
\textbf{Test Font}       & \textbf{C}   & \textbf{A} & \textbf{T} & \textbf{S} \\ \hline
\textbf{Proposed Method} & \textbf{100} & 31.11      & 16.67      & 18.89      \\ \hline
\end{tabular}
\end{table}

\subsection{Variations of Proposed CNN Architecture}
\label{subsec:DeepPhone_CNNarch}
We evaluate the effect of average and max pooling operations and various activation functions.
We compare commonly used activation functions, including exponential linear unit (ELU), rectified linear unit (ReLU), parametric rectified linear unit (PReLU), and hyperbolic tangent (Tanh).
The results in~Table~\ref{tab:DeepPhone_pool_activation} show that the proposed method provides similar performance with all combinations.
However, a combination of Tanh activation function and average pooling performs slightly better than others providing 98.81\% average classification accuracy.

\begin{table}[t!]
    \centering
	\caption{Comparison of variations in CNN architecture: classification accuracy statistics and standard deviation in brackets using a 5 $\times$ 2 cross-validation approach.}
	\label{tab:DeepPhone_pool_activation}
	\resizebox{0.5\textwidth}{!}{
\begin{tabular}{|c|c|c|c|c|c|c|}
\hline
\multirow{3}{*}{\textbf{\begin{tabular}[c]{@{}c@{}}Activation \\ Function\end{tabular}}} & \multicolumn{6}{c|}{\textbf{Pooling Function}}                                                                                                                                                                                                                                                          \\ \cline{2-7} 
                                                                                         & \multicolumn{3}{c|}{\textbf{Max}}                                                                                                                  & \multicolumn{3}{c|}{\textbf{Avg}}                                                                                                                  \\ \cline{2-7} 
                                                                                         & \textbf{\begin{tabular}[c]{@{}c@{}}Mean\\ (in \%)\end{tabular}} & \textbf{\begin{tabular}[c]{@{}c@{}}Median\\ (in \%)\end{tabular}} & \textbf{Std} & \textbf{\begin{tabular}[c]{@{}c@{}}Mean\\ (in \%)\end{tabular}} & \textbf{\begin{tabular}[c]{@{}c@{}}Median\\ (in \%)\end{tabular}} & \textbf{Std} \\ \hline
\textbf{ELU}                                                                             & 98.24                                                           & 98.15                                                             & 0.58         & 98.37                                                           & 98.43                                                             & 0.84         \\ \hline
\textbf{ReLU}                                                                            & 98.42                                                           & 98.29                                                             & 0.66         & 98.41                                                           & 98.29                                                             & 0.64         \\ \hline
\textbf{PReLU}                                                                           & 98.46                                                           & 98.29                                                             & \textbf{0.60}         & 98.37                                                           & 98.29                                                             & 0.69         \\ \hline
\textbf{Tanh}                                                                            & 98.72                                                           & \textbf{98.66}                                                             & 0.61         & \textbf{98.81}                                                           & 98.61                                                             & 0.66         \\ \hline
\end{tabular}
}
\end{table}

\section{Conclusion}
\label{sec:DeepPhone_Conclusion}
We proposed an approach to identify the source printer of a printed document via its camera acquired image.
The use of smartphones achieves significant letter and document level classification accuracy that would be useful in many practical scenarios as they provide more flexibility to end-users.
Our method uses a two-channel combination of the native letter image and its noise residual image to learn a single CNN model.
We use a series of experiments to show the efficacy of the proposed method on a newly created dataset.
Experiments on the scanner dataset in~\cite{joshiIcassp2018}~(Table 1) highlighted a consistent improvement
due to a combination of native and noise-residual images.
In contrast, experiments on our camera acquired document dataset revealed that the noise-residual image does not have a significant impact on the performance.
One possible reason for this could be that the printer-specific noise residual estimation from camera acquired data is already corrupted by unwanted noise due to camera artifacts and blur.
Nonetheless, the proposed method performs better than state-of-the-art methods, including a popular state-of-the-art image classification method (i.e., DenseNet~\cite{DenseCNN2017}) providing an average classification accuracy of 98.42 \% using a 5 $\times$ 2 cross-validation protocol.

We also evaluated the performance of the proposed method under varying image acquisition scenarios, including the effect of angle between the camera and document planes, and intensity of indoor lighting.
Results indicate that the performance is almost consistent across all variations considered in this work, except down tilt scenario.
The analysis revealed that increased perspective distortion degraded letter extraction performance with the down-tilt dataset.
The proposed method achieves an average letter-level classification accuracy of 90.33\% by learning a single CNN model from about 1 million letter images consisting of all types of letters printed in a specific font type and size.
Such a high accuracy highlights the strong ability of the proposed method to extract printer-specific signatures from camera-acquired letter images and paves the way for certain types of intra-page forgery detection.
Also, 98.01\% of average page-level accuracy is obtained, which lowers the dependence on sufficient availability of a specific letter type.

Similar to previous data-driven approaches, the proposed method fails under the cross-font scenario.
Toner level may impact printer signature, so in real-world applications, it is expected that the training data is available over the entire printer toner cycle.
Several other scenarios may impact a printer's signature, including printer age and characteristics of printing paper.

Future work may include a large scale evaluation of source printer attribution using camera acquired document images in the wild.
Also, the impact of other variations in text like font size and other languages may be explored.


	
\section{Acknowledgment}
This material is based upon work partially supported by a grant from the Department of Science and Technology (DST), New Delhi, India, under Award Number ECR/2015/000583 and Visvesvaraya PhD Scheme, Ministry of the Electronics \& Information Technology (MeitY), Government of India. 
Any opinions, findings, and conclusions or recommendations expressed in this material are those of the author(s) and do not necessarily reflect the views of the funding agencies.

\ifCLASSOPTIONcaptionsoff
  \newpage
\fi

\bibliographystyle{IEEEtran}
\bibliography{Deep_Phone}

\begin{thebibliography}{10}
\providecommand{\url}[1]{#1}
\csname url@samestyle\endcsname
\providecommand{\newblock}{\relax}
\providecommand{\bibinfo}[2]{#2}
\providecommand{\BIBentrySTDinterwordspacing}{\spaceskip=0pt\relax}
\providecommand{\BIBentryALTinterwordstretchfactor}{4}
\providecommand{\BIBentryALTinterwordspacing}{\spaceskip=\fontdimen2\font plus
\BIBentryALTinterwordstretchfactor\fontdimen3\font minus
  \fontdimen4\font\relax}
\providecommand{\BIBforeignlanguage}[2]{{%
\expandafter\ifx\csname l@#1\endcsname\relax
\typeout{** WARNING: IEEEtran.bst: No hyphenation pattern has been}%
\typeout{** loaded for the language `#1'. Using the pattern for}%
\typeout{** the default language instead.}%
\else
\language=\csname l@#1\endcsname
\fi
#2}}
\providecommand{\BIBdecl}{\relax}
\BIBdecl

\bibitem{web:paperusage2018}
\BIBentryALTinterwordspacing
Global forest products facts and figures 2018. Accessed: Feb. 27, 2020.
  [Online]. Available: \url{http://www.fao.org/3/ca7415en/CA7415EN.pdf}
\BIBentrySTDinterwordspacing

\bibitem{chiang2009printer}
P.-J. Chiang, N.~Khanna, A.~K. Mikkilineni, M.~V.~O. Segovia, S.~Suh, J.~P.
  Allebach, G.~T.-C. Chiu, and E.~J. Delp, ``{Printer And Scanner Forensics},''
  \emph{IEEE Signal Process. Mag.}, vol.~26, no.~2, pp. 72--83, 2009.

\bibitem{ferreira2017data}
A.~Ferreira, L.~Bondi, L.~Baroffio, P.~Bestagini, J.~Huang, J.~dos Santos,
  S.~Tubaro, and A.~Rocha, ``{Data-Driven Feature Characterization Techniques
  for Laser Printer Attribution},'' \emph{IEEE Trans. Inf. Forensics Security},
  vol.~12, no.~8, pp. 1860--1873, 2017.

\bibitem{Chiang2011}
P.-J. Chiang, J.~P. Allebach, and G.~T.~C. Chiu, ``{Extrinsic Signature
  Embedding And Detection In Electrophotographic Halftoned Images Through
  Exposure Modulation},'' \emph{IEEE Trans. Inf. Forensics Security}, vol.~6,
  no.~3, pp. 946--959, 2011.

\bibitem{mikkilineni2011forensic}
A.~K. Mikkilineni, N.~Khanna, and E.~J. Delp, ``{Forensic Printer Detection
  Using Intrinsic Signatures.}'' in \emph{Media Forensics and Security}, 2011,
  p. 78800R.

\bibitem{burie2015icdar2015}
J.-C. Burie, J.~Chazalon, M.~Coustaty, S.~Eskenazi, M.~M. Luqman, M.~Mehri,
  N.~Nayef, J.-M. Ogier, S.~Prum, and M.~Rusi{\~n}ol, ``Icdar2015 competition
  on smartphone document capture and ocr (smartdoc),'' in \emph{2015 13th
  International Conference on Document Analysis and Recognition (ICDAR)}.\hskip
  1em plus 0.5em minus 0.4em\relax IEEE, 2015, pp. 1161--1165.

\bibitem{kim2019learning}
D.-G. Kim, J.-U. Hou, and H.-K. Lee, ``Learning deep features for source color
  laser printer identification based on cascaded learning,''
  \emph{Neurocomputing}, vol. 365, pp. 219--228, 2019.

\bibitem{ferreira2015laser}
A.~Ferreira, L.~C. Navarro, G.~Pinheiro, J.~A. dos Santos, and A.~Rocha,
  ``{Laser Printer Attribution: Exploring New Features And Beyond},''
  \emph{Forensic Science International}, vol. 247, pp. 105--125, 2015.

\bibitem{joshi2018single}
S.~Joshi and N.~Khanna, ``{Single Classifier-based Passive System for Source
  Printer Classification Using Local Texture Features},'' \emph{IEEE
  Transactions on Information Forensics and Security}, vol.~13, no.~7, pp.
  1603--1614, 2018.

\bibitem{joshi2019source}
------, ``Source printer classification using printer specific local texture
  descriptor,'' \emph{IEEE Transactions on Information Forensics and Security},
  vol.~15, pp. 160--171, 2019.

\bibitem{joshiIcassp2018}
S.~Joshi, M.~Lamba, V.~Goyal, and N.~Khanna, ``{Augmented Data And Improved
  Noise Residual-Based CNN For Printer Source Identification},'' \emph{IEEE
  International Conference on Accoustics, Speech and Signal Processing
  (ICASSP)}, 2018, {(To Appear)}.

\bibitem{girard2013criminalistics}
J.~E. Girard, \emph{{Criminalistics: Forensic Science, Crime, And Terrorism}},
  3rd~ed.\hskip 1em plus 0.5em minus 0.4em\relax Burlington, MA: Jones \&
  Bartlett Publishers, 2013.

\bibitem{oliver2002use}
J.~Oliver and J.~Chen, ``{Use of Signature Analysis to Discriminate Digital
  Printing Technologies},'' in \emph{Proc. IS\&T's NIP18: Int. Conf. Digital
  Printing Technologies}, San Diego, CA, Sep. 2002, pp. 218--222.

\bibitem{Lampert2007}
C.~H. Lampert, L.~Mei, and T.~M. Breuel, ``{Printing Technique Classification
  For Document Counterfeit Detection},'' in \emph{Proc. Int. Conf.
  Computational Intelligence and Security}, vol.~1, Guangzhou, China, Nov.
  2006, pp. 639--644.

\bibitem{schulze2008evaluation}
C.~Schulze, M.~Schreyer, A.~Stahl, and T.~M. Breuel, ``{Evaluation Of
  Graylevel-features For Printing Technique Classification In High-throughput
  Document Management Systems},'' in \emph{Proc. 2nd Int. Workshop on
  Computational Forensics (IWCF)}.\hskip 1em plus 0.5em minus 0.4em\relax
  Washington, DC: Springer, Aug. 2008, pp. 35--46.

\bibitem{Schulze2009}
C.~Schulze, M.~Schreyer, A.~Stahl, and T.~Breuel, ``{Using Dct Features for
  Printing Technique and Copy Detection},'' \emph{Advances in Digital Forensics
  V}, pp. 95--106, 2009.

\bibitem{schreyer2009intelligent}
M.~Schreyer, C.~Schulze, A.~Stahl, and W.~Effelsberg, ``{Intelligent Printing
  Technique Recognition And Photocopy Detection For Forensic Document
  Examination.}'' in \emph{Proc. Informatiktage: Fachwissenschaftlicher
  Informatik-Kongress 27. und 28}, vol.~8, Bonn, Germany, Mar. 2009, pp.
  39--42.

\bibitem{roy2010authentication}
A.~Roy, B.~Halder, and U.~Garain, ``{Authentication of Currency Notes Through
  Printing Technique Verification},'' in \emph{Proc. Seventh Indian Conf. on
  Comput. Vision, Graphics and Image Process. (ICVGIP)}.\hskip 1em plus 0.5em
  minus 0.4em\relax Chennai, India: ACM, 2010, pp. 383--390.

\bibitem{shang2014detecting}
S.~Shang, N.~Memon, and X.~Kong, ``{Detecting Documents Forged By Printing And
  Copying},'' \emph{EURASIP Journal on Advances in Signal Process.}, vol. 2014,
  no.~1, pp. 1--13, 2014.

\bibitem{Mikkilineni2005}
A.~K. Mikkilineni, P.-J. Chiang, G.~N. Ali, G.~T.~C. Chiu, J.~P. Allebach, and
  E.~J. Delp, ``{Printer Identification Based On Graylevel Co-occurrence
  Features For Security And Forensic Applications},'' in \emph{Proc. SPIE Int.
  Conf. on Security, Steganography, and Watermarking of Multimedia Contents
  VII}, vol. 5681, San Jose, CA, March 2005, pp. 430--441.

\bibitem{mikkilineni2005printer}
A.~K. Mikkilineni, O.~Arslan, P.-J. Chiang, R.~M. Kumontoy, J.~P. Allebach,
  G.~T.-C. Chiu, and E.~J. Delp, ``{Printer Forensics Using SVM Techniques},''
  in \emph{NIP \& Digital Fabrication Conf.}, vol. 2005, no.~1, 2005, pp.
  223--226.

\bibitem{Tsai2014}
M.~J. Tsai, J.~S. Yin, I.~Yuadi, and J.~Liu, ``{Digital Forensics of Printed
  Source Identification for Chinese Characters},'' \emph{Multimedia Tools and
  Applications}, vol.~73, no.~3, pp. 2129--2155, Dec 2014.

\bibitem{tsai2018decision}
M.-J. Tsai, I.~Yuadi, and Y.-H. Tao, ``{Decision-theoretic Model to Identify
  Printed Sources},'' \emph{Multimedia Tools and Applications}, pp. 1--45,
  2018.

\bibitem{Elkasrawi2014}
S.~Elkasrawi and F.~Shafait, ``{Printer Identification Using Supervised
  Learning for Document Forgery Detection},'' in \emph{Proc. 11th IAPR Int.
  Workshop on Document Analysis Systems (DAS)}, France, Apr. 2014, pp.
  146--150.

\bibitem{tsai2015japanese}
M.-J. Tsai, C.-L. Hsu, J.-S. Yin, and I.~Yuadi, ``{Japanese character based
  printed source identification},'' in \emph{Proc. IEEE Int. Symposium on
  Circuits and Systems (ISCAS)}, Lisbon, Portugal, May 2015, pp. 2800--2803.

\bibitem{he2015spatial}
K.~He, X.~Zhang, S.~Ren, and J.~Sun, ``{Spatial Pyramid Pooling in Deep
  Convolutional Networks for Visual Recognition},'' \emph{IEEE Transactions on
  Pattern Analysis and Machine Intelligence}, vol.~37, no.~9, pp. 1904--1916,
  2015.

\bibitem{kee2008printer}
E.~Kee and H.~Farid, ``{Printer Profiling For Forensics And Ballistics},'' in
  \emph{Proc. 10th ACM Workshop Multimedia and Security}, Oxford, United
  Kingdom, Sep. 2008, pp. 3--10.

\bibitem{zhou2015text}
Q.~Zhou, Y.~Yan, T.~Fang, X.~Luo, and Q.~Chen, ``{Text-Independent Printer
  Identification Based On Texture Synthesis},'' \emph{Multimedia Tools and
  Applications}, vol.~75, no.~10, pp. 5557--5580, May 2016.

\bibitem{bulan2009geometric}
O.~Bulan, J.~Mao, and G.~Sharma, ``{Geometric Distortion Signatures For Printer
  Identification},'' in \emph{Proc. IEEE Int. Conf. Acoustics, Speech and
  Signal Process.}, Taipei, Taiwan, Apr. 2009, pp. 1401--1404.

\bibitem{wu2009printer}
Y.~Wu, X.~Kong, X.~You, and Y.~Guo, ``{Printer Forensics Based on Page
  Document's Geometric Distortion},'' in \emph{Proc. IEEE International
  Conference on Image Processing (ICIP)}.\hskip 1em plus 0.5em minus
  0.4em\relax IEEE, 2009, pp. 2909--2912.

\bibitem{jain2019passive}
H.~Jain, S.~Joshi, G.~Gupta, and N.~Khanna, ``Passive classification of source
  printer using text-line-level geometric distortion signatures from scanned
  images of printed documents,'' \emph{Multimedia Tools and Applications}, pp.
  1--24, 2019.

\bibitem{krizhevsky2012imagenet}
A.~Krizhevsky, I.~Sutskever, and G.~E. Hinton, ``{Imagenet Classification with
  Deep Convolutional Neural Networks},'' in \emph{Proc. Advances in Neural
  Information Processing Systems}, 2012, pp. 1097--1105.

\bibitem{szegedy2015going}
C.~Szegedy, W.~Liu, Y.~Jia, P.~Sermanet, S.~Reed, D.~Anguelov, D.~Erhan,
  V.~Vanhoucke, and A.~Rabinovich, ``{Going Deeper with Convolutions},'' in
  \emph{Proceedings of the IEEE Conference on Computer Vision and Pattern
  Recognition}, 2015, pp. 1--9.

\bibitem{ioffe2015batch}
S.~Ioffe and C.~Szegedy, ``{Batch Normalization: Accelerating Deep Network
  Training by Reducing Internal Covariate Shift},'' in \emph{Proceedings of the
  32nd International Conference on International Conference on Machine Learning
  - Volume 37}, ser. ICML’15.\hskip 1em plus 0.5em minus 0.4em\relax
  JMLR.org, 2015, p. 448–456.

\bibitem{fisher2001digital}
F.~Fisher, ``Digital camera for document acquisition,'' in \emph{Proc.
  symposium on document image understanding technology}, 2001, pp. 75--83.

\bibitem{liang2005camera}
J.~Liang, D.~Doermann, and H.~Li, ``Camera-based analysis of text and
  documents: a survey,'' \emph{International Journal of Document Analysis and
  Recognition (IJDAR)}, vol.~7, no. 2-3, pp. 84--104, 2005.

\bibitem{MemonBook2013}
M.~A. Arbib, \emph{Digital Image Forensics: There is More to a Picture than
  Meets the Eye}.\hskip 1em plus 0.5em minus 0.4em\relax USA, NY, New York:
  Springer, 2013.

\bibitem{kingma2014adam}
D.~P. Kingma and J.~Ba, ``Adam: A method for stochastic optimization,''
  \emph{arXiv preprint arXiv:1412.6980}, 2014.

\bibitem{DenseCNN2017}
G.~{Huang}, Z.~{Liu}, L.~v.~d. {Maaten}, and K.~Q. {Weinberger}, ``{Densely
  Connected Convolutional Networks},'' in \emph{2017 IEEE Conference on
  Computer Vision and Pattern Recognition (CVPR)}, July 2017, pp. 2261--2269.

\bibitem{kerasapplication}
\BIBentryALTinterwordspacing
{{DenseNet} from keras team}. Accessed: Jan. 3, 2020. [Online]. Available:
  \url{https://github.com/keras-team/keras-applications/blob/master/keras_applications/densenet.py}
\BIBentrySTDinterwordspacing

\bibitem{van2013text}
J.~Van~Beusekom, F.~Shafait, and T.~M. Breuel, ``{Text-line Examination for
  Document Forgery Detection},'' \emph{International Journal on Document
  Analysis and Recognition (IJDAR)}, vol.~16, no.~2, pp. 189--207, 2013.

\end{thebibliography}
\end{document}